\documentclass[runningheads]{llncs}
\usepackage[T1]{fontenc}
\usepackage{graphicx}
\usepackage{xcolor}
\usepackage{authblk}
 \usepackage{subfigure}
 \usepackage[font=scriptsize,labelfont=bf]{caption}
\begin{document}
\title{A Search for Prompts: Generating Structured Answers from Contracts}
%
%
\author{Adam  Roegiest \and
Radha Chitta\and
Jonathan Donnelly\and 
Maya Lash \and 
Alexandra Vtyurina \and 
François Longtin
}
\authorrunning{Roegiest et. al.}
\institute{
Zuva Inc., Toronto, Ontario, Canada \\
\email{\{adam,radha.chitta,jonny.donnelly,maya,sasha, francois\}@zuva.ai}
}



\newcommand{\red}[1]{\textcolor{red}{#1}}
\maketitle

\begin{abstract}
In many legal processes being able to action on the concrete implication
of a legal question can be valuable to automating human review or signalling
certain conditions (e.g., alerts around automatic renewal).
To support such tasks, we present a form of legal question answering that 
seeks to return one (or more) fixed answers for a question about a contract clause.
After showing that unstructured generative question answering can have questionable
outcomes for such a task, 
we discuss our exploration methodology for legal question answering prompts using OpenAI's \textit{GPT-3.5-Turbo} and provide a summary
of insights.

Using insights gleaned from our qualitative experiences,
we compare our proposed template prompts against a common semantic matching approach
and find that our prompt templates are far more accurate despite being less 
reliable in the exact response return.
With some additional tweaks to prompts and the use of in-context learning,
we are able to further improve the performance of our proposed strategy while
maximizing the reliability of responses as best we can.

\keywords{Structured answers  \and Legal Question Answering\and Generative AI \and Prompt Engineering.}
\end{abstract}
\section{Introduction}
Understanding obligations and restrictions in contracts is as useful to lawyers as it is to lay-people\footnote{Especially for lay-people, it is probably more useful as a means to avoid the costs of dealing with a lawyer.} and 
goes beyond just finding relevant information in a contract to understanding the underlying
implications (e.g., whether a tenant can sublet an apartment).
On the basis of a single contract, this is not too tedious to perform manually when relevant
clauses are extracted for the reader but quickly becomes onerous and time-consuming at scale
or when legal expertise is not a strong suit.
This understanding of not just the clauses but the implications present in contracts can allow
individuals and organizations to make smarter decisions and have more confidence when doing so.
This is especially paramount when certain implications can have potentially very negative effects
(e.g., unlimited liability guarantees, lowest price guarantees) depending on which side of the implication one is on.

With the increasing popularity of Generative AI and applications to ``chat'' with one's documents (e.g., ChatPDF\footnote{\url{https://www.chatpdf.com/}}, docGPT\footnote{\url{https://github.com/cesarhuret/docGPT}}, PDFGPT.IO\footnote{\url{https://pdfgpt.io/}}, CaseText\footnote{\url{https://casetext.com/}}),
there is the potential to more easily facilitate this understanding of legal obligations and restrictions
at scale. Inspired by this, we sought to investigate how we might leverage Generative AI, 
specifically OpenAI's GPT-3.5-Turbo, 
to help us generate structured answers to questions about clauses in legal documents. 
Taking inspiration from closed question-answering~\cite{dimitrakis2020survey} and textual entailment~\cite{yin2019benchmarking},
our task is such that we are provided a set of contextual clauses about
which a user would like to ask questions with predefined outputs (e.g., multiple choice)
rather than a free-form answer.
Such outputs can facilitate the creation of business workflows that can easily take action on the results of the questions
without resorting to complex understanding of a generated answer or manual human review.
While such outputs are perhaps less compelling than having a ``conversation'' with a document, 
we believe that this approach scales better to a document collection and workflow automation.
Moreover, we show that more natural prompts can very easily exhibit consistency
and reliability issues when examining the responses generated in a ``chat'' interaction paradigm (Section \ref{sec:open-ended-generation}).

While prompt engineering is a new area of research~\cite{pengfei2023pre,reynolds2021prompt,zhou2022large,white2023prompt}
and book publication\footnote{With over 500 English language books dedicated to the topic on Amazon as of April 29, 2022.},
it is a task that neither we nor the users of our product actually desire to do.
Our end goal then is to find and use prompts that provide consistent answers for the same clause and reliable answers across clauses.
In particular, we are interested in finding reusable templates that allow end users to ``fill in answer options'' and not worry about
overall prompt engineering. 
This allows end users to customize and modify options to their needs but it also results in system builders not needing to foresee all possible information needs and build bespoke prompts for each of them.
To find such templates, four team members tested over 700 prompt and clause combinations to discover critical aspects
of prompts and prompt templates that we detail in Section~\ref{sec:template-discovery}.
Using the insights gleaned, we tested our prompt templates in a more rigorous evaluation in Section~\ref{sec:sag}, where we compare our approach
to more traditional semantic matching techniques and find that our prompts can outperform such techniques.
We conclude with a discussion of limitations and future work. 

\section{Background}
Generative models such as LaMDA~\footnote{https://blog.google/technology/ai/lamda/}, LLaMA~\footnote{https://ai.facebook.com/blog/large-language-model-llama-meta-ai/}, and GPT-n series~\footnote{https://platform.openai.com/docs/models} are self-supervised models which learn to predict the next token in a sequence of tokens~\cite{radford2018improving,touvron2023llama}. 
These \textit{large} language models have billions to trillions of parameters and are pre-trained on massive corpora of texts. 
Unlike the early pre-trained language models, they require little to no fine-tuning and perform well in zero-shot and few-shot settings. 

In a zero-shot generation task, the model is provided with the inputs and an instruction in natural language, and the response of the model can be parsed to obtain the answer. 
For open-domain QA the only input is the question but for closed-domain the input also includes the context.
In a few-shot setting, the model is also provided with a few examples, which have the indirect effect of ``fine-tuning'' the model~\cite{brown2020language}. 
Recent large language models (e.g.\textit{GPT-3.5-Turbo}, Vicuna-13B, Alpaca) have been fine-tuned, either using supervised or reinforcement learning, with prompts for a diverse set of natural language tasks~\cite{sanh2021multitask,ouyang2022training,wang2022benchmarking,alpaca,vicuna2023}. They have generally outperformed earlier language models on most natural language generation benchmarks including question answering, summarization, and translation~\cite{zhang2023benchmarking}. Instruction-tuned models achieve better performance than fine-tuned models but 
 can suffer from two major issues~\cite{ji2023survey}: 
\vspace{-0.5em}
\begin{itemize}
    \item Inconsistency: Results generated have been found to be overly reliant on the phrasing of the prompt (e.g., small prompt changes producing inexplicable differences in generated responses, multiple runs yielding different results)
    \item Hallucination: Models were found to generate false/irrelevant information in some instances, which introduces risk when accuracy is important.
\end{itemize}
\vspace{-0.5em}
Prompt engineering techniques~\cite{gao-etal-2021-making,pengfei2023pre,white2023prompt} aim to design prompts that mitigate these issues. For instance, in~\cite{gao-etal-2021-making} a smaller language model, such as T5, is tuned to "auto-complete" the prompts for the large language model.  Prompt templates are created in~\cite{pengfei2023pre} and~\cite{white2023prompt} which can be auto-filled. While these methods have had some success, this research is still ongoing.
In our work, we use an OpenAI provided large language model, \textit{GPT-3.5-Turbo}, and determine if it is possible to create prompt templates that can be used to consistently answer questions based on contract clauses.

\section{Open-Ended Response Generation}
\label{sec:open-ended-generation}
In many legal processes, lawyers look to take an action on a particular answer option rather than a summary or the raw text of a clause itself. For example, while dealing with  environmental indemnification, a lawyer (representing a landlord)
may wish to know in what contracts the tenants indemnify the landlord rather than any explicit conditions around the indemnification (which may be relevant later).
By identifying which contracts do not have such a clause, the lawyer may have particular advice for the landlord. 
When only indemnification clauses are presented, however, the lawyer would be required to read them to answer these questions (e.g., direction of indemnification).
An ideal system would produce a concise and consistent answer when asked ``who indemnifies whom?'' with respect to the clause.

\begin{figure}
    \centering
    \includegraphics[width=0.7\textwidth]{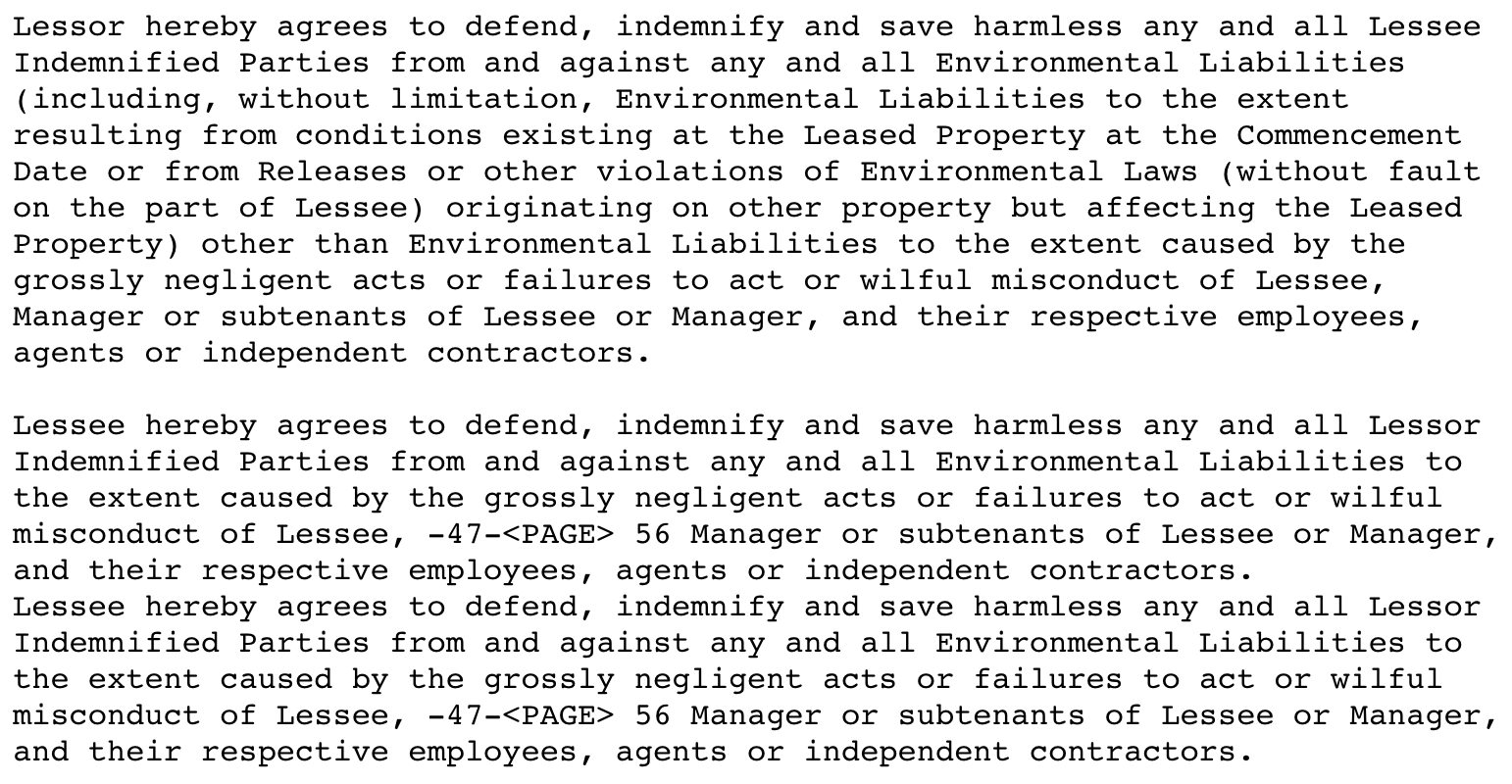}
    \caption{An example of an environmental indemnity clause with mutual indemnification.}
     \label{fig:indemnity}
\end{figure}

Figure \ref{fig:indemnity} is an example indemnity clause from an arbitrary lease from our collection of legal documents. 
As an example, if one prompts OpenAI's GPT-3.5-Turbo model (with temperature set to 0) with the following: \textit{In the clause below, who indemnifies whom?} and then append the example clause, we get the following responses after regenerating a handful of times (with varying repetitions):
\begin{itemize}
\scriptsize
\item The Lessor indemnifies the Lessee Indemnified Parties, and the Lessee indemnifies the Lessor Indemnified Parties.
\item Lessor indemnifies Lessee Indemnified Parties for any and all Environmental Liabilities, except those caused by the grossly negligent or wilful misconduct of Lessee, Manager, or subtenants of Lessee or Manager, and their respective employees, agents or independent contractors. Lessee indemnifies Lessor Indemnified Parties for Environmental Liabilities caused by the grossly negligent or wilful misconduct of Lessee, Manager, or subtenants of Lessee or Manager, and their respective employees, agents or independent contractors.
\item Lessor indemnifies Lessee Indemnified Parties. Lessee indemnifies Lessor Indemnified Parties.
\end{itemize}

\noindent While the first and third are close, they aren't identical and the second is more of an overall summary. 
We note that these are all correct responses but also not something one could plugin in to a workflow automation tool without a lot of additional post-processing (e.g., in the flavour of nugget matching~\cite{dang-etal-2006-trec}).
To further illustrate this problem, we present the responses from the same prompt with a different mutual environmental indemnity clause:
\begin{itemize}
    \scriptsize
    \item Tenant indemnifies Landlord for claims arising from the introduction of hazardous material on or after the Effective Date, and Landlord indemnifies Tenant for claims arising from the existence of hazardous material or violation of Environmental Requirements prior to the Effective Date.
    \item In the first paragraph Tenant indemnifies and defends Landlord, and in the second paragraph Landlord indemnifies and defends Tenant.
    \item Tenant indemnifies Landlord. Landlord indemnifies Tenant.
\end{itemize}

\noindent Again, we see responses that are factually correct but are also quite different from the responses for the previous clause.
As should be apparent by now, this inconsistency and unreliability as well as response verbosity is not ideal for simplifying work process
and may just exacerbate things in the long run.

\section{Finding Reusable Prompt Templates}
\label{sec:template-discovery}
Before the launch of OpenAI's ChatGPT in December 2022, 
we had spent time investigating the utility of MNLI~\cite{adina-2018-mnli} pre-trained encoder models\footnote{Specifically \url{https://huggingface.co/valhalla/distilbart-mnli-12-1} and related models.} for structured question answering (described in Section~\ref{sec:entailment})
but found that the technique was only suitable when clauses and options could fit in relatively small context windows of such models~\cite{veeranna2016using,yin2019benchmarking}.
Using similar clauses to the previous section but modelling the question closer to one of entailment, we were able to achieve a more desirable result
with ChatGPT on a small set of examples.
From this success, we built a tool\footnote{Description of the tool is omitted here for brevity. It is available at \url{https://github.com/zuvaai/gpt-tool}.} that would allow internal experts to try various prompt and option combinations using previously annotated legal clauses as the context for the prompt.
The remainder of this section discusses the overall methodology used and the key findings from that process on how prompt templates might be structured.

\subsection{Prompt Exploration Methodology}

Over a 3.5 week period, several members of our team tried over 700 prompt and clause combinations to explore how sensitive prompts were to variations in clauses.\footnote{We note that for all practical purposes, outside of some initial testing, temperature was set fairly consistently to 0.}
Prompts were based on an existing set of questions that we curated ourselves but also includes some end-user provided examples.
All of the prompt contexts were previously annotated for supervised machine learning tasks for the purposes of content extraction, previously described in ~\cite{roegiest2018,donnelly2020} and derived from legal documents primarily from EDGAR (\url{https://www.sec.gov/edgar}) and SEDAR (\url{https://www.sedar.com/}).
While using the tool, a team member would select a clause type (e.g., assignment, environmental indemnity) and craft a prompt to answer a related question.
They would then run this prompt on as many examples of this clause type as they desired, rating the responses, and making tweaks as they went along to improve the consistency and reliability of responses.
We note that some tweaks were made to the tool's interface during this period with the goal of simplifying the testing and evaluation experience with no material effect made to the outcomes discussed.


Team members were free to progress at their own rate and develop their own strategies to avoid locking anyone into a particular style.
Though active discussions took place around different prompting styles and how certain edge cases could be avoided or addressed more generally.
Invariably this approach has meant that some redundant work has likely occurred which is not ideal but was unlikely to be avoided.
This also means that different understandings of the various prompting strategies have also developed which we believe is worth the additional effort.

\subsection{Prompt Template Key Aspects}
\label{sec:qual}

Using the logs from our tool, one of the authors conducted a very light ``thematic analysis'' inspired review of the logs to examine
trends in both prompts and in the notes themselves. 
While we have not ``cracked the code'' of one-size-fits-all prompt templates and suspect that doing so is a Sisyphean task,
we have learned some interesting things on how to steer a generative AI model in returning answers in the structured format
that we find desirable.
In the following, we provide the insights and key aspects we have gleaned from the analysis of logs and group discussions 
that occurred while undertaking this search for prompt templates.

\subsubsection{Justification}
As their name implies, generative models are designed to generate text and when answering questions one such source 
of generation is in justifying a choice.
Conceptually this is useful but as a component in workflow/process automation, it is not.
Unlike other approaches~\cite{yu2022legal},
we very consistently added a constraint instructing the model to respond ``\textit{without providing justification}'' or
``\textit{without providing other information}'' to nudge the model away from generating text that may or may not be
reliable and further complicate downstream processing.

\subsubsection{Option Selection}
An additional aspect of our prompts was in specifying responses should \textit{only} be from a predefined list of options which we found helps to guide the model to actually select one of the options.
Unfortunately, this had a confounding factor in that depending on how this was specified (e.g., \textit{``only select the numbered option that is implied by the clause''}),
the generative model may respond with just the number, ``Option \textit{i}'', or the formatted option (e.g., $(3)$, $3.$), or the entire option, or just the answer text of the option (e.g., ``Landlord indemnifies Tenant'').
None of these responses are bad but does mean that some downstream processing is necessary to clean-up responses to a canonical format but could be resolved by setting \textit{max\_tokens} when generating a response.

\subsubsection{Options over Prompts}

The overarching structure of a prompt can certainly have an effect on how succinct and accurate a response can be and is
why we have devoted so much time to finding good templates.
But throughout the process we noticed that how answers were phrased had more impact on the end response generated by the model.
In particular, we found that options should be as complete a thought or sentence as possible to avoid leaving openings for the model to begin generating additional text (e.g., often elaborating on an option to finish the sentence). 
We also saw examples where ending options in full-stops or wrapping them in quotations could mitigate generation even with 
sentence fragments but this was not consistent.

\subsubsection{Escape Hatches}

One of the most interesting aspects of our early trials and even subsequent more thorough investigations with our prompt tool is the ability of the model to make assumptions when lacking sufficient data or context in a clause.
As an example, in the  case of environmental indemnities where the parties might actually be named (e.g., SmithCo, Joe Smith) rather than a defined term (e.g., landlord, lessee). 
In such a case, the model often just made assumptions about which party fulfilled a specific role without any prompting or evidence.
Interestingly, the model managed to get it right a fair number of times but we hesitate to attribute this to any particular reason.
Such aspects can be remediated by ensuring those details are also provided as context but the willingness of the model to make
assumptions is not ideal.

To combat this and prevent the model from making assumptions or trying to free-form respond with an ``unable to determine''-style response,
we tried to provide additional ``rules'' or ``escape hatches'' in our prompts to allow the model to default to a \textit{safe} choice.
The most generic rule that we used was: \textit{If you cannot determine which of the conditions are implied, respond with the exact text: `The clause is silent.'}
This statement required a bit of fine-tuning since an initial attempt to use ``answer with ...''  often resulted in the model
then generating a response along the desired lines but the exact phrase.
We posit that this is due to the nature of how the model was trained and less due to explicit phrasing (i.e., prompts asking
the model to answer would often be generative in nature in the training set).
We have also tried similar but less generic escape hatches where needed (e.g., the aforementioned indemnity situation)
and the model will make use of them with sufficient wordsmithing.

\subsubsection{Multiple Selections}

Having a model produce multiple selections is critical when it is possible for multiple conditions to be 
true and specifying all combinations is not reasonably feasible (i.e., more than a handful of combinations).
While we did not spend quite as much time on these types of prompts as others,
we did determine that it is possible to do this and that there might be several possible ways to perform the task.

The most obvious is just to direct the model to choose multiple options 
(e.g., \textit{``If multiple options are implied by the clause, select all of the numbered options that apply.''}) 
but this has a surprising potential implication. 
The model can be ``too smart'' and perhaps rely too much on a non-legal understanding of English and conflate multiple
terms of art in the legal world. 
For example, prior to refinement of options, the model would often conflate ``requiring to give notice'' and
``requiring consent'' and would tend to default towards the latter when determining obligations around the assignment of a contract to another party. 

An alternative approach that we found had potential is to frame each option as a sub-prompt that required the 
model to answer ``yes'' or ``no'' (as to whether the clause supported the option).
In our basic exploration, this often felt easier to get working more quickly but we have yet to expend
a lot of effort exploring further.
The downside is that the aforementioned problems around assumptions and language conflation were more pronounced
and a bit harder to tackle.
Moreover, it is not clear whether or not just iteratively asking the options as individual prompts (in the style of \cite{yu2022legal}) either independently
or as part of a larger ``conversation'' would yield superior results.

\section{Structured Answer Generation} 
\label{sec:sag}
Before we address our evaluation of structured generation,
we first discuss some context setting for choosing this avenue of research.
It would natural to believe that training a discriminative multi-class classifier to obtain structured answers
to the questions presented in this work is a reasonable first choice.
Such solutions (e.g., SVMs, logistic regression) are well understood and highly optimized but come with a 
flaw, they require sufficient examples of each class to be effective (i.e., low prevalence classes can be hard to overcome).
For example, landlord indemnification is a relatively rare occurrence in our data and one for which data augmentation is
not guaranteed to work.
Indeed, we have attempted to address this in prior work~\cite{chitta2019reliable}, 
it did not consistently work well for especially low prevalence or nuanced differences between options.
It has been our experience that such cases arise naturally in the legal domain where rare outcomes tend to be the most problematic
and equally hard to identify.

Inspired by recent uses of large language models and textual entailment to facilitate ``zero-shot'' question-answering, we put such
methods to the test.
We begin with an examination of semantic matching as a baseline before comparing with generative model
prompting techniques and the differences in option choices.
We then explore the consistency and reliability of our empirical results before providing some  initial experiments
detailing how in-context learning~\cite{min2022rethinking,xie2022explanation} can improve even bad prompts. 
\subsection{Semantic Similarity and Answer Prediction}\label{sec:entailment}
\begin{table}[]
    \centering
    \resizebox{0.99\textwidth}{!}{
    \begin{tabular}{|c||c|c|c|c|c||c|c|c|c|c||c|c|c|c|c||c|c|c|c|c|}
    \hline
    &\multicolumn{5}{c||}{\textbf{S1}} & \multicolumn{5}{c||}{\textbf{S2}} &\multicolumn{5}{c||}{\textbf{S3}} &\multicolumn{5}{c|}{\textbf{S4}} \\
    \cline{2-21}
    &\textbf{1 (6)} &  \textbf{2 (71)}& \textbf{3 (39)}& \textbf{4 (5)}  &  \textbf{A} &\textbf{1 (6)} &  \textbf{2 (71)}& \textbf{3 (39)}& \textbf{4 (5)}  &  \textbf{A} &\textbf{1 (6)} &  \textbf{2 (71)}& \textbf{3 (39)}& \textbf{4 (5)}  &  \textbf{A} &\textbf{1 (6)} &  \textbf{2 (71)}& \textbf{3 (39)}& \textbf{4 (5)}  &  \textbf{A} \\
    \hline
    \textbf{PIL-LegalBert} &0 &10 &34 &0 &0.36& 0& 30 & 0  &0 &0.25 & 0& 4& 11& 2& 0.14&0 & 23& 3 & 3& 0.24\\
    \hline
    \textbf{OpenAI Ada}&0 &1 &0 &3 &0.03 &0 & 0& 0&4 &0.03& 0& 1& 2& 2& 0.04& 0& 0& 20&2 &0.18\\
    \hline
    \textbf{OpenAI GPT-3.5 Turbo}&6 &60 &23 &1 &0.74& 6 &44 &27 & 2& 0.65 &5 &58 &24 &1 & 0.73&5 &67 &6 &1 &0.65 \\
    \hline
    \textbf{OpenAI GPT-4} & 5 &66 & 39 & 1& 0.91 & 6 & 66 & 39 & 0 & 0.92& 5 & 66 & 39 &0 & 0.91 & 5 &67  & 39 & 1 & 0.92\\
    \hline
    \textbf{PaLM 2}&  0 & 58 & 39 & 1 &  0.84 &4 &53  & 39& 1 & 0.79 & 4&55& 39&0 &  0.81&4 & 56 & 39 & 0 & 0.82 \\
    \hline
      \end{tabular}}
    \caption{  \scriptsize Number of correct predictions for each answer option and accuracy (A) with different phrasings of the answer options and the prompts using the zero-shot semantic similarity approach (rows 1-2) and generation (rows 3-5) for the indemnity question. The true distribution of the answers is listed along with the answer options in brackets.}
    \label{tab:entailment-results-indemnity}
         \vspace{-3em}
\end{table}

\begin{table}[htbp]
    \centering
    \resizebox{0.75\textwidth}{!}{
    \begin{tabular}
    {|c||c|c|c|c|c|c||c|c|c|c|c|c|}
    \hline
         &\multicolumn{6}{c||}{\textbf{T1}} & \multicolumn{6}{c|}{\textbf{T2}} \\
         \cline{2-13}
         &\textbf{1 (4)} &  \textbf{2 (81)}& \textbf{3 (40)}& \textbf{4 (2)} & \textbf{5 (57)} &   \textbf{A}&\textbf{1 (4)} &  \textbf{2 (81)}& \textbf{3 (40)}& \textbf{4 (2)} & \textbf{5 (57)} &   \textbf{A} \\
         \hline
         \textbf{PIL-LegalBert} &0 &11 &22 &0&0&0.23& 0& 4&1 &0&21&0.18 \\
         \hline
         \textbf{OpenAI Ada}&0& 0& 0& 0&6&0.04 &0&0 &0 &0 &1&0.01 \\
         \hline
         \textbf{OpenAI GPT-3.5 Turbo}& 3& 56& 19& 1&0&0.55 &0 &66 & 18& 0&1&0.59 \\
         \hline
         \textbf{OpenAI GPT-4}& 1& 21 & 33 &1 & 2& 0.43& 2& 41&16 &0 & 0& 0.48 \\
         \hline
         \textbf{PaLM2}&0 &65 &26 & 0& 1& 0.65& 0& 60& 17&0&24  &0.71\\
        \hline 
     \end{tabular}}
    \caption{\scriptsize Number of correct predictions for each answer option and accuracy (A) with different phrasings of the answer options and the prompts using the zero-shot semantic similarity approach (rows 1-2) and generation (rows 3-5) for the information-sharing question. The true distribution of the answers is listed along with the answer option in brackets.}
    \label{tab:entailment-results-infosharing}
    \vspace{-3em}
\end{table}     
Given that our work was inspired by natural language inference (Section \ref{sec:open-ended-generation}), 
we investigate the utility of a baseline method that is not generative but is less restricted by the minimal context length
in many popular encoder models (e.g., BERT and its variants).
We focus on zero-shot question answering using semantic similarity, which aims to identify which of the options from a given set of options is best supported by the given text~\cite{veeranna2016using,yin2019benchmarking}. 

Given the text of the clause \(T\), and a set of answer options \(H\), we use a sequence embedding model \({M}\) to obtain the embeddings for the clause  and the answer options and then predict the answer which has the highest similarity with the clause as follows: 
$\arg\max_{h \in H} \cos \left( M \left( T\right), M\left(h \right) \right)$.

\begin{figure}
\centering
\begin{minipage}{.45\textwidth}
    \centering
    \includegraphics[width=\columnwidth]{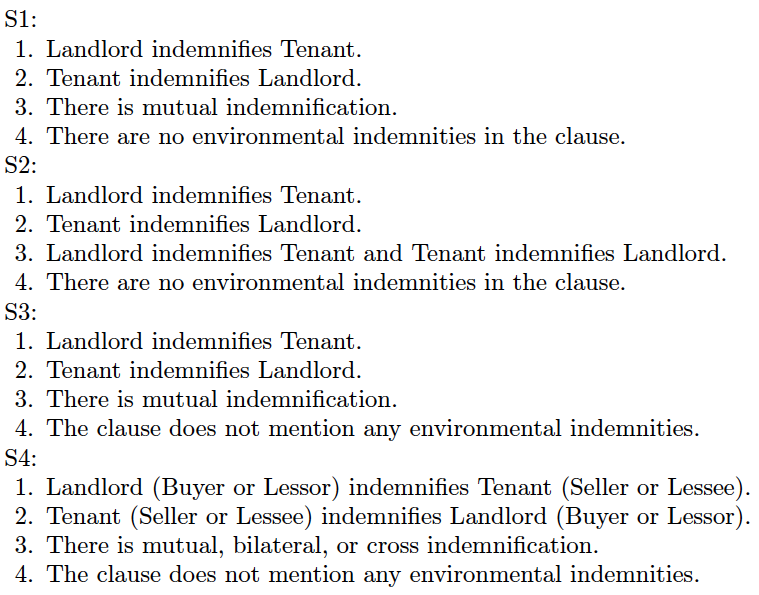}
    \caption{Four sets of answer options for the indemnity question.}
     \label{fig:indemnity_options}
\end{minipage}\hfill
\begin{minipage}{.45\textwidth}
    \centering
    \includegraphics[width=\columnwidth]{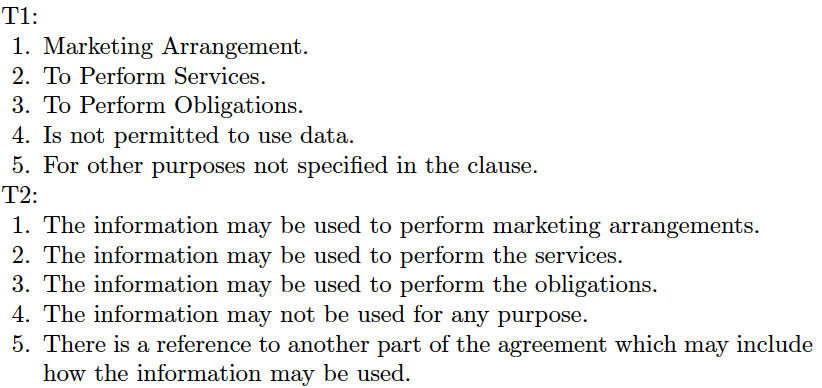}
    \caption{Two sets of answer options for the information-sharing question.}
     \label{fig:info_sharing_options}
\end{minipage}
\end{figure}

We use two legal questions \textit{"In the clause below, who indemnifies whom?"} (``indemnity question'') and \textit{"For what purpose are the parties sharing information according to the clause below?"} (``information-sharing question'') to test the effectiveness of this method. 
The test datasets for the two questions consisted of a set of legal clauses, which we assume are previously extracted from the contracts through manual means detailed in prior work~\cite{roegiest2018}. 
For the indemnity question, we used a set of 121 \textit{"environmental indemnity"} clauses, which describe the provisions for security or protection from losses or damage caused by environmental contamination or disasters. 
For the information-sharing question, our test dataset comprised of 143 \textit{"Permitted Use of Confidential Information"} clauses, which describe the purposes for which information shared between the parties may be used. 
From our observations in Section~\ref{sec:open-ended-generation} that the phrasing of the question and the answer options plays an important role in the answers generated by the models, we tested out four sets of answer options (S1- S4) for the indemnity question and two sets of answer options (T1-T2) for the information-sharing question, shown in Figures~\ref{fig:indemnity_options} and~\ref{fig:info_sharing_options} respectively.

The test clauses for the indemnity question have the distribution of the answers: 6, 71, 39, and 5 for each of the four options. 
Unlike the indemnity question where only one of the options is applicable, the information-sharing question is a \textit{multi-select} question (i.e., multiple options may be applicable). The test clauses for this question have the answers distributed as 4, 81, 40, 2, and 57, respectively. There are also 13 clauses that do not contain sufficient information to determine the correct answer.
For the information-sharing question, we perform a lenient evaluation of the results and consider the prediction correct if at least one of the correct options was predicted.
Such lenient evaluation is to make overall comparisons between methods easier without having to also attempt to learn (or guess) a threshold in the case of semantic similarity.

We ensured all clauses were less than \(20,000\) characters in length in order to adhere to the token limits of the OpenAPI models.
As we can see, both datasets exhibit the prevalence problem we described at the beginning of this section,
which means that traditional classifiers are unlikely to handle this scenario well~\cite{japkowicz2000class,Chawla_2002,chitta2019reliable}.

We used two large language models to compare the performance of this method: (1) the LegalBERT model trained on the Pile-Of-Law dataset~\cite{hendersonkrass2022pileoflaw}\footnote{\url{https://huggingface.co/pile-of-law/legalbert-large-1.7M-2}} and (2) the OpenAI Ada model~\cite{neelakantan2022text}.

For a given model, we used cosine similarity between options and clause embeddings to choose the most similar option to a given clause.
We then computed the accuracy of option to ground truth for each embedding model and present the results of this evaluation 
in Tables~\ref{tab:entailment-results-indemnity} and~\ref{tab:entailment-results-infosharing}. 
We observe that the LegalBERT embeddings, which were trained on contracts, performed moderately better than the OpenAPI Ada embeddings, for which the exact training details are not known.
For instance, the terms Lessee and Tenant have a 90\% similarity using LegalBERT embeddings, whereas the similarity drops to 87\% when using the OpenAI Ada embeddings. 
Though this may not seem like a substantive difference, 
it is apparent that explicitly including the terms Lessee and Tenant in the answer options \(S4\) makes a moderate difference in the performance using the Ada embeddings but less so for LegalBERT. 
Unsurprisingly, phrasing the answer options differently leads to different accuracy regardless of model choice. 
For example, to identify mutual indemnification, the option ``\textit{There is mutual indemnification}'' in \(S1\) and \(S3\) leads to a better result than any other form of the answer option. 

\subsection{Generating Structured Answers} \label{sec:generative-ai}

From our manual exploration for reusable prompts, we had sufficient evidence to indicate that we should more thoroughly
and empirically investigate this approach for structured answer generation.
Accordingly, we evaluate the effectiveness of our approach on the two questions and examine how our approach
holds up against the semantic similarity approach of the previous section.
To test the effectiveness of structured answer generation, we employed OpenAI's \textit{GPT-3.5-Turbo}~\cite{ouyang2022training} and \textit{GPT-4}~\cite{openai2023gpt4}, and Google's \textit{PaLM2}~\cite{anil2023palm} Generative AI. \newline

\noindent We started with the following prompt:
\begin{scriptsize}
\begin{verbatim}
P1:  Referring only to the information contained in the clause below, only select which one 
     of the below numbered options is implied by the clause, without providing any other 
     information or justification.  If you cannot determine which of the conditions are 
     implied, respond with the exact text: “The clause is silent”.
    {{Options}}
    {{Clause}}
\end{verbatim}
\end{scriptsize}
\noindent \(\lbrace \lbrace Options \rbrace \rbrace \) is replaced by the options in each set from the previous section and \(\lbrace \lbrace Clause \rbrace \rbrace\) is replaced by the text of the clause. 
A prompt styled in this manner aims to ensure that only \textbf{one} of the answer options is generated, 
and that the option generated is not followed by any additional text, 
thereby rendering a structured response.
The prompt provides an escape hatch for the model (returning ``The clause is silent.'') if all the relevant answer options were not provided or the clause does not contain sufficient information. 
The response from the API is parsed to obtain the structured answer to the question. 

From Tables~\ref{tab:entailment-results-indemnity} and~\ref{tab:entailment-results-infosharing}, we can easily see that the performance of this approach non-trivially and substantially improves over the semantic matching approach.
Consistent with the similarity results,
different forms of answer options perform differently even for generative models but with still generally okay performance
in the worst case.
Interestingly, there does not appear to be any strong correlation in how well answer options will perform between either
approach (e.g., \textit{S3} is the second best performing for \textit{GPT-3.5-Turbo} but the worst for semantic similarity).
We also note that the rare clauses where either the ``Landlord indemnifies Tenant'' or there is ``No indemnification'' are correctly answered by the generative models. Options \textit{T2} yield substantive improvement over options \textit{T1} showing that clear, complete sentences as options lead to better accuracy. Among the generative models, the more complex \textit{GPT-4} and \textit{PaLM2} obviously perform better than \textit{GPT-3.5-Turbo}. On the indemnity question, both these models are able to correctly identify all instances of mutual indemnification. However, they don't do as well on the information-sharing question, most likely because the prompts and the options are not well-structured. A prompt styled for one model, despite performing relatively well, is not guaranteed to yield optimal results  with other generative models. While not surprising, it does suggest a developer ought to choose a model and stick with it. 

Despite styling the prompt to ensure that only one of the answer options was generated, 
we found that there was some lack of consistency in the responses (particularly those of \textit{GPT-3.5-Turbo}) across clauses. 
For instance, when provided with options \textbf{S1} for the indemnity question, 
the response, ``\textit{The clause implies that Tenant indemnifies Landlord.}'', was generated for some clauses instead of ``\textit{Tenant indemnifies Landlord.}''. 
The model also generated ``\textit{Lessee indemnifies Lessor}'' for some clauses instead of ``\textit{Tenant indemnifies Landlord.}'',
which we posit in due to the nature of the language used in some clauses (i.e., the model tries to match the language of the clause).
In two to three instances, the model also provided a summary of the clause in addition to the answer option. 
For example, one of the responses generated was ``\textit{Lessee indemnifies Lessor and its Affiliates for any and all Environmental Costs incurred in connection with the presence or alleged presence of Hazardous Substances or Mold on or near the Leased Property.}'' 

This necessitated some post-processing (e.g., substring matching) of the response to determine which answer option applied where possible.
Such an outcome is not what we desired to have happen but is an aspect we are willing to bend on, especially when such post-processing
is not complex (i.e., substring or synonym substitution is preferable to nugget-matching).
 
There were 16, 15, 12, and 9 unique responses generated by \textit{GPT-3.5-Turbo} for each of the four answer options for the indemnity question. It is our best guess that \textbf{S4} generates the least number of unique responses because the answer options explicitly note that the terms Lessee, Tenant, and Seller and the terms Lessor, Landlord, and Buyer represent the same entities. 
Surprisingly, the answers generated for the information-sharing question were more consistent. 
There were only two instances where a summary of the clause was generated.   

While this approach yielded unintended response creativity by the generative model,
we think this is a strong indicator that such an approach may be viable for more use cases. 
To further attempt to refine our approach and prompt phrasings,
we detail in the next section how ``prompt sensitive'' a generative model can be
and how we could limit some of the post-processing.
For brevity and more focused analysis, we present results only for \textit{GPT-3.5-Turbo} henceforth.

\subsection{Sensitivity to Prompts}

As we saw in the previous section, 
one glaring issue with using a generative model for structured answer generation is that it can still get creative in its
responses despite our attempts to limit this creativity.
Our first idea was that perhaps the order of options would play a part but shuffling the order of options did not change
any outcomes and so we do not report on them further.
Instead, we report on how changing aspects of the prompt template affect the outcome of the results by introducing new rules
or adding more explicit rules for the model to follow.

To address the issues around the model getting too creative in its responses when using \textbf{P1},
we sought to limit the results by requiring the model to respond only with the numbered option rather than
the entire option:
\begin{scriptsize}
\begin{verbatim}
P2: Referring only to the information contained in the clause below, only select the numbered 
    option that is implied by the clause, without providing any other information or 
    justification. If you cannot determine which of the conditions are implied, respond with 
    the exact text: “The clause is silent”.
    {{Options}}
    {{Clause}}
\end{verbatim}    
\end{scriptsize}

We also created a prompt, which would allow the model to make multiple selections and making the rules and options more explicit:
\begin{scriptsize}
\begin{verbatim}
P3: Using the text provided, follow the subsequent instructions:
    {{Clause}}
    Respond with all options which are implied by the provided text, without providing any 
    other information or justification and by following the rules.
    Rules:
        - If it cannot be determined which of the conditions are implied or if it is required 
        to make assumptions, respond with the exact text: "Unable to determine."
        - If the terms in the options are not used in the text, respond with the exact text: 
        "Unable to determine."
    Options:
    {{Options}}
\end{verbatim}    
\end{scriptsize}

Table~\ref{tab:generation_prompts_indemnity} shows the difference in the number of correct answers generated for the indemnity question using
our three prompt templates and the answer options \textbf{S1}.
Perhaps unsurprisingly, the small modifications made to \textbf{P2} do not substantively change the model's ability to
produce the correct answer nor did it particularly limit the creativity of responses generated by the model.
The much larger changes in \textbf{P3}, on the other hand, brought about substantive improvements to the model's responses
but also meant that we had slightly more post-processing to do to accommodate multiple responses being returned.
While the latter was not unexpected, it does illustrate a further complication that could spur more model creativity
in other situations, especially those where the answer options may be sentence fragments.

For the information-sharing question, we found that asking the model to pretend to be a party to the agreement, explicitly including the question in the prompt, and allowing for multiple option selection improved the accuracy substantially, as seen in Table~\ref{tab:generation_prompts_infosharing}. This type of prompt styling does not lend as easily to the creation of templates, as it requires the user to provide a question along with the options. However, the improvement in accuracy is worth the additional effort. 
\begin{scriptsize}
\begin{verbatim}
P4: Read the following permitted use of confidential information legal clause: 
    {{Clause}}
    Pretend you are a party to the agreement in which the permitted use of confidential 
    information legal clause you have read exists in. You only know what you have read in 
    this prompt. For what purposes are you allowed to use the confidential information? If 
    the clause does not specify the purpose for which you may use the confidential  
    information, respond with: "Unable to determine". In your response, only include the  
    following most correct groups:
    {{Options}}
    In your response, only include the bucket names above. Do not provide an explanation 
    or additional information.
\end{verbatim}
\vspace{-2em}
\end{scriptsize}

\begin{table}
\centering
\begin{minipage}{.45\textwidth}
      \scriptsize
      \centering
    \begin{tabular}{|c||c|c|c|c||c|}
    \hline
    & \textbf{1 (6)} & \textbf{2 (71)} & \textbf{3 (39)} & \textbf{4 (5)} & \textbf{A}\\
    \hline
         \textbf{P1} &6  & 60 & 23& 1&0.74  \\
        \hline
        \textbf{P2} & 5 & 65 &19&1  & 0.71 \\
        \hline
        \textbf{P3}& 4 & 62  &35 & 1& 0.85 \\
        \hline
    \end{tabular}
    \caption{  \scriptsize Number of correct predictions for each answer option and accuracy (A) with different prompts and options S1 for the indemnity question. The true distribution of the answers is listed along with the answer options in brackets. }
    \label{tab:generation_prompts_indemnity}
\end{minipage}\hfill
\begin{minipage}{.45\textwidth}
    \centering
    \resizebox{\textwidth}{!}{
    \begin{tabular}{|c||c|c|c|c|c||c|}
    \hline
    & \textbf{1 (4)} & \textbf{2 (81)}& \textbf{3 (40)} & \textbf{4 (2)}  & \textbf{5 (57)}   & \textbf{A} \\
    \hline
        \textbf{P1}& 0 & 66 & 18 & 0 &1 & 0.59  \\
        \hline
        \textbf{P4}& 4 & 73 &24 &2  & 5 & 0.64\\
        \hline
    \end{tabular}}
    \caption{  \scriptsize Number of correct predictions for each answer option and accuracy (A) with different prompts and options T2 for the information-sharing question. The true distribution of the answers is listed along with the answer options in brackets.}
    \label{tab:generation_prompts_infosharing}
\end{minipage}
\vspace{-3em}
\end{table}


Throughout our investigation, 
we have regularly found that using a consistent template and aiming for option correctness has tended to produce 
more viable results in the longer term but does have some periods of inconsistency and unreliability in the results
until the right options are found.
Indeed, trying to min-max the prompt template may not always yield the most optimal outcomes as \textbf{P2} is ever
so slightly worse than \textbf{P1} in our tests.
While \textbf{P3} and \textbf{P4} did substantially improve the overall effectiveness, this improvement was based on a decent amount of trial and error in our prompt tool and is not an approach we want to undertake for every possible legal question,
especially if it risks the model becoming creative in its responses.
Nor is this an approach that we want an end-user to have to replicate as it would be far too onerous for them.

All that being said,
in our tests throughout this section and the previous one, we have found that once a prompt template and a set
of options have hit a steady state in consistency then they do, in fact, stay consistent (i.e., regenerating the response for the same combination of clause, prompt, and options produces the same result). 
Using \textbf{P1}, we used the \textit{GPT-3.5-Turbo} API to re-run the generation of responses 5 times and saw that
the results were extremely consistent with no more than a couple of responses changing.
Unfortunately, the reliability of responses is not ideal as we discussed earlier but
does show that the right combination of prompt and options can non-trivially reduce the amount of variation
(e.g., P4 having slightly more than half the variations than P1).

\subsection{In-Context Learning}
If we set aside issues with the model generating the exact response we want and focus solely on the effectiveness of the technique,
we see that our structured answer generation approach is definitely better than using semantic similarity (Section~\ref{sec:generative-ai}) but that does not mean the technique is ready for end-user consumption (i.e., business processes likely want better than 0.74 accuracy).
A simple way to potentially help the model ``understand'' the task that we are instructing it to perform is 
to use some form of few-shot learning, specifically in-context learning~\cite{min2022rethinking,xie2022explanation}, which has been shown to improve model effectiveness~\cite{brown2020language,gao-etal-2021-making}.

In our case, we leverage the fact that \textit{GPT-3.5-Turbo} is a chat-style model and can ``seed'' it with labelled
examples as part of the conversation (i.e., we pre-populate earlier parts of the conversation with both prompt and answer).
To do so, we augment a simple  prompt, \textbf{P1}, and answer set, \textbf{S1}, with randomly selected human labelled examples for each of the four options. 
We create two example sets for the indemnity question, \textbf{E1} and \textbf{E2}, each containing $4$ randomly selected example clauses, one for each answer option. 
As seen in Table \ref{tab:fewshot}, there is reasonable improvement in the accuracy with both example sets. 
But when augmented with both example sets (two examples for each answer option), there is a stark improvement in accuracy. 
We find that 
the model is able to identify tricky cases like mutual indemnification when previously it had struggled with just \textbf{P1} and \textbf{S1}.
Moreover, this technique can dramatically reduce the amount of post-processing necessary to map responses back to options.
For example, for the indemnity question, the number of responses needing cleanup drop by at least 50 percent; 83 responses with \textbf{S1} to 22 responses with \textbf{S1} and the two example sets.

\begin{table}[]
\vspace{-1em}
\centering
  \scriptsize
    \begin{tabular}{|c||c|c|c|c|}
    \hline
        & \textbf{No examples} & \textbf{E1} & \textbf{E2} & \textbf{E1 + E2}  \\
        \hline
        \hline
       Indemnity &0.74&0.83 &0.79  & 0.87\\
        \hline
        Information-sharing & 0.55 & 0.61 & 0.62 & 0.64 \\
        \hline
    \end{tabular}
    \caption{  \scriptsize Accuracy when \textit{GPT-3.5-Turbo} is provided in-context with different sets of examples, prompt \textbf{P1}, and answer sets \textbf{S1} and \textbf{T1} for the indemnity and information-sharing questions.}
    \label{tab:fewshot}
   \vspace{-3em}
\end{table}

Utilizing in-context learning has allowed us to improve response generation substantively while incurring the small expense of annotating some
additional clauses for the appropriate answer.
This indicates that there may be a valuable trade-off to identify between prompt (or option) engineering and just spending
the time providing a small handful of examples.
Exploring methods to determine when prompt engineering has diminishing results and it would be better to provide examples
is an interesting avenue of future work that we plan to explore. 

\section{Limitations} \label{sec:limitations}


For the sake of experimental simplicity and general ease of use, we have focused most of our testing on  the \textit{GPT-3.5-Turbo} model. Preliminary results on \textit{GPT-4} and \textit{PaLM2} in Tables~\ref{tab:entailment-results-indemnity} and~\ref{tab:entailment-results-infosharing} show that the prompts that work well with \textit{GPT-3.5-Turbo} can be strong starting points but may not be optimal out of the gate. 
Based on this, we feel comfortable saying that our proposed solution is viable on other generative AI models with the
reasonable caveat that no one prompt is a golden ticket for success.
Moreover, as there are increasing numbers of models being made available for commercial use~\cite{dolly,pythia,falcon,llama2},
we focused on \textit{GPT-3.5-Turbo} because of its reasonable costs and general widespread adoption. 


We have also largely operated under the assumption that the user of the underlying system has the correct context
for a clause available to them to ask questions about.
While the end-to-end effectiveness will matter in a final system (e.g., using some other information retrieval systems to first
identify the clauses), 
we believe that our prompt structure and escape hatches will help a model overcome false positives but leave investigation to future work.

\section{Conclusions and Future Work}\label{sec:conclusions}
We have presented a structured answer generation task based on identifying the correct answer for a legal question given
an associated clause from a document, detailed some issues with relying solely on natural language question answering,
and then present insights gleaned from a manual exploration of the prompt space over 3.5 weeks.
Using these insights, we compared the reliability and
consistency of structured answer generation for two legal questions and several hundred examples
and found that our proposed technique is superior to semantic matching between clauses and answer options,
especially through small tweaks to prompts and the use of in-context learning.

Looking forward, we plan to more exhaustively investigate how many different
prompt templates might be needed for our use cases and whether or not we can identify good and bad answer options 
(i.e., to avoid users having to guess and check).
Additionally, being able to automatically detect whether or not further refinement of options or providing
examples for in-context learning is a lucrative avenue as it may reduce user frustration to just provide labelled examples.
As the generative model space stabilizes, we also plan to investigate the transferability of our prompts to other
models to determine their actual reusability.
 \bibliographystyle{splncs04}
 \bibliography{sample-base}

\end{document}